\documentclass[sigconf, nonacm, authorversion]{acmart}

\AtBeginDocument{%
  }

\usepackage{amsmath}
\usepackage{mathtools}
\usepackage{amsthm}
\usepackage{booktabs}
\usepackage{multirow} 
\usepackage{algorithm}
\usepackage{algorithmic}

\begin{document}

\title{THeGAU: Type-Aware Heterogeneous Graph Autoencoder and Augmentation}

\author{Ming-Yi Hong}
\affiliation{%
    \department{Data Science Degree Program}
    \institution{National Taiwan University and Academia Sinica}
    \city{Taipei}
    \country{Taiwan}
}

\author{Miao-Chen Chiang}
\affiliation{%
    \department{Data Science Degree Program}
    \institution{National Taiwan University and Academia Sinica}
    \city{Taipei}
    \country{Taiwan}
}

\author{Youchen Teng}
\affiliation{%
    \department{Graduate Institute of Communication Engineering}
    \institution{National Taiwan University}
    \city{Taipei}
    \country{Taiwan}
}

\author{Yu-Hsiang Wang}
\affiliation{%
    \institution{University of Illinois Urbana-Champaign}
    \department{Electrical \& Computer Engineering}
    \state{Illinois}
    \country{United States}
}

\author{Chih-Yu Wang}
\affiliation{%
    \department{Research Center for Information Technology Innovation}
    \institution{Academia Sinica}
    \city{Taipei}
    \country{Taiwan}
}

\author{Che Lin}
\authornote{Corresponding author: chelin@ntu.edu.tw}
\affiliation{%
    \department{Department of Electrical Engineering}
    \institution{National Taiwan University}
    \city{Taipei}
    \country{Taiwan}
}

\renewcommand{\shortauthors}{Ming-Yi Hong et al.}

\begin{abstract}
Heterogeneous Graph Neural Networks (HGNNs) are effective for modeling Heterogeneous Information Networks (HINs), which encode complex multi-typed entities and relations. However, HGNNs often suffer from type information loss and structural noise, limiting their representational fidelity and generalization. We propose THeGAU, a model-agnostic framework that combines a type-aware graph autoencoder with guided graph augmentation to improve node classification. THeGAU reconstructs schema-valid edges as an auxiliary task to preserve node-type semantics and introduces a decoder-driven augmentation mechanism to selectively refine noisy structures. This joint design enhances robustness, accuracy, and efficiency while significantly reducing computational overhead. Extensive experiments on three benchmark HIN datasets---IMDB, ACM, and DBLP---demonstrate that THeGAU consistently outperforms existing HGNN methods, achieving state-of-the-art performance across multiple backbones.
\end{abstract}




\maketitle

\section{Introduction}

In recent years, graph neural networks (GNNs) have become a critical technology in graph mining research for graph-structured data representation learning, with broad applications in molecular design, recommendation systems, computer vision, and graph mining \cite{stokes2020deep, jin2018learning, wu2019session, 10.1145/3535101, shi2020point, li2019graph, He2022}. Researchers have extended GNNs from homogeneous to heterogeneous, forming heterogeneous information networks (HINs) to better capture complex data relationships in real-world scenarios. These networks comprise diverse node and edge types, each carrying distinct information and characteristics, presenting unique challenges due to their inherent complexity.

To improve the handling of complex and varied data structures in HINs, heterogeneous graph neural networks (HGNNs) have made significant strides in addressing the complexity of HINs. Meta-path-based models \shortcite{Wang2019} \shortcite{Fu2020} rely on predefined meta-paths for information propagation. Transformer-based models like HGT \cite{Yun2019GTN, Hu2020HGT} leverage transformer architectures to identify key message-passing paths without predefined meta-paths. Building on these approaches, SimpleHGN \cite{Lv2021SimpleHGN} introduced a shared feature space design that simplifies computations by mapping diverse node attributes into a unified space, improving accuracy. However, merging information from different node types into a unified space can lead to over-squeezing, limiting effective feature extraction.
To address this, SlotGAT \cite{slotgat} introduces slot-based message passing to maintain semantic distinctions across feature spaces, while TreeXGNN \cite{10096251} employs a tree-based feature extractor to enhance feature identification prior to GNN integration. Despite their effectiveness, TreeXGNN’s two-stage design adds complexity to parameter tuning for new datasets. 
Relying solely on an HGNN encoder may result in the irreversible loss of essential node and edge-type information during message passing. Loss of such information could be critical for follow-up tasks. Unfortunately, most existing semi-supervised HGNN algorithms lack explicit mechanisms to preserve this structural heterogeneity throughout the learning process. These challenges highlight the difficulties HGNNs face in learning robust representations and achieving consistent performance.

Semi-supervised encoder-decoder learning approaches have been proposed in the literature to guarantee the preservation of critical information in the learning process. It employs end-to-end training strategies that incorporate auxiliary losses to streamline model learning and improve performance, making the design and integration of these losses a pivotal area of research. 
Such an approach has achieved notable success in fields such as computer vision \cite{Kingma2013AutoEncodingVB, JimenezRezende2014StochasticBA} and GNNs \cite{Kipf2016VariationalGA, dalvi2022variational}. However, existing works focus on specific data formats, such as images and homogeneous graphs, leaving a clear gap in the development of semi-supervised learning frameworks for HGNNs that incorporate auxiliary tasks to learn general representations and achieve stable results.

Furthermore, HINs often exhibit substantial structural noise, such as missing, irrelevant, or redundant edges, which compromise graph integrity and hinder effective message propagation. Such imperfections can significantly degrade the performance of HGNNs---particularly in semi-supervised settings. While recent advances have enhanced heterogeneous graph representations through pre-training, contrastive learning, prompt tuning, and interpretable message passing \cite{shgp, heco2021, hetgpt, hgprompt, iehgcn}, the challenge of mitigating structural noise---especially in the form of spurious or redundant connections---remains largely underexplored. This highlights a critical gap: the need for effective, model-compatible graph augmentation techniques that explicitly denoise HINs to improve both representational fidelity and downstream predictive performance.

To address these challenges, we propose \textbf{Type-aware Heterogeneous Graph Autoencoder and AUgmentation} (THeGAU), a novel and model-agnostic framework that enhances the generalization and predictive capabilities of HGNNs through autoencoders and heterogeneous graph augmentation. Our main contributions are summarized as follows: %\footnote{Open-source code will be released upon acceptance.}
\begin{itemize}
\item 
We introduce the THeGAU framework with type-aware graph autoencoders specifically designed for HIN to enable end-to-end semi-supervised training and heterogeneous graph augmentation. To the best of our knowledge, this is the first framework to accomplish this for HINs. 
\item 
The proposed type-aware graph autoencoders serve as an auxiliary task to reconstruct typed edges, preserve structural information, stabilize training, and prevent overfitting, thereby achieving better model generalization.
Additionally, the skip-layer design enhances shared-space training in HGNNs, improving feature extraction and reducing over-smoothing and information loss. 
\item 
The proposed type-aware graph autoencoder enables graph augmentation by removing noise and adding relevant edges, enhancing the signal-to-noise ratio within the graph, and simultaneously reducing both storage and computational complexity for edge prediction. This design achieves state-of-the-art classification performance on benchmark datasets.
\end{itemize}

\section{Related Works}\label{sec:RelatedWorks}

\subsection{Heterogeneous Graph Neural Networks}

HGNNs are tailored for handling HINs. Early models like Heterogeneous Graph Attention Network (HAN) \cite{Wang2019} and the Meta Path Aggregated Graph Neural Network (MAGNN) \cite{Fu2020} rely on predefined meta-paths, which require domain expertise and limit scalability. Graph Transformer Network (GTN) \cite{Yun2019GTN} addresses this by learning meta-paths automatically. Heterogeneous Graph Transformer (HGT) \cite{Hu2020HGT} further advances HGNNs by employing a transformer-based architecture and subgraph sampling, enabling efficient learning on large and complex heterogeneous graphs.

SimpleHGN \cite{Lv2021SimpleHGN} utilizes projection layers to project information to the shared feature space and aggregates information with an edge-type attention mechanism. While this design reduces model complexity, it overlooks the inherent characteristics of HINs, precisely the diversity of node and edge types. The projection process results in a loss of type-specific attributes, thereby compromising the model's ability to retain crucial type information. 
Therefore, SlotGAT \cite{slotgat} introduced a slot-based GNN message-passing approach to preserve semantic distinctions across feature spaces, while TreeXGNN \cite{10096251} employs a tree-based feature extractor before the HGNN to refine type-specific information. While these methods aid feature extraction, they struggle with mixed-type information and structural complexities in HINs, highlighting the need for more expressive and robust HGNN frameworks.

\subsection{Graph AutoEncoders}

Variational Graph Autoencoders (VGAE) \cite{Kipf2016VariationalGA} was originally proposed to leverage a Variational Autoencoder for self-supervised link prediction on homogeneous graphs. Later extensions adapted VGAE to heterogeneous settings \cite{dalvi2022variational}, but their focus on link prediction limits applicability to broader tasks such as node classification.
Further exploring autoencoders in HINs, the Heterogeneous Graph Attention Auto-Encoder (HGATE) \cite{9664297} uses a meta-path-based model but underperforms in transductive node classification \cite{Lv2021SimpleHGN, slotgat, 10096251}. Similarly, HGMAE \cite{HGMAE23}, which employs meta-path-based masked graph encoders, struggles to achieve high prediction accuracy.
The challenge primarily stems from the unstructured form of HINs compared to those in natural language processing (NLP) and computer science, which complicates the application of graph representation learning. Graph autoencoder techniques have been successfully applied in NLP. For instance, the Graph Attention Transformer Encoder (GATE) \cite{Ahmad2020GATEGA} integrates encoder–decoder architectures for tasks like cross-lingual relation and event extraction. However, its use remains specific to NLP applications. In contrast, our work aims to develop a comprehensive heterogeneous graph decoder framework that learns from the inherent information of HINs, thereby integrating various state-of-the-art HGNN encoders. 

\subsection{Graph Data Augmentation}

Data augmentation is widely used in NLP and computer vision (e.g., synonym replacement \cite{Zhang2015CharacterlevelCN}, text variants \cite{kafle-etal-2017-data}, back-translation methods \cite{sennrich-etal-2016-improving, Xie2019UnsupervisedDA, edunov-etal-2018-understanding}, and image transformations \cite{Shorten2019ASO}), but these methods are not directly applicable to HINs, whose non-Euclidean structure, lack of a fixed grid, and unordered node connections present distinct challenges. 

Several techniques have been developed to address the complexities of homogeneous graphs. DropEdge \cite{Rong2020DropEdge} improves model robustness by randomly eliminating a fraction of graph edges before their introduction into GNNs. AdaEdge \cite{Chen2019MeasuringAR} modifies the graph structure through a two-stage process, adjusting edges based on a confidence measure to enhance the integrity of information within the graph. Furthermore, GAUG \cite{Zhao2020DataAF} utilizes edge predictors to optimize the transmission of useful information and reduce noise within the graph, achieving promising results. 
While many recent studies \cite{gmixup2022, lagnn2022, Zhang_Zhu_Song_Koniusz_King_2023, Zhou_Gong_2023, NEURIPS2024_d450dcee} still focus on homogeneous graphs, our work bridges the gap by extending these augmentation strategies to heterogeneous graphs, developing tailored designs to leverage their unique structures, and improving model performance across diverse applications. 

\begin{figure*}[t]
\centering
\includegraphics[width=\textwidth]{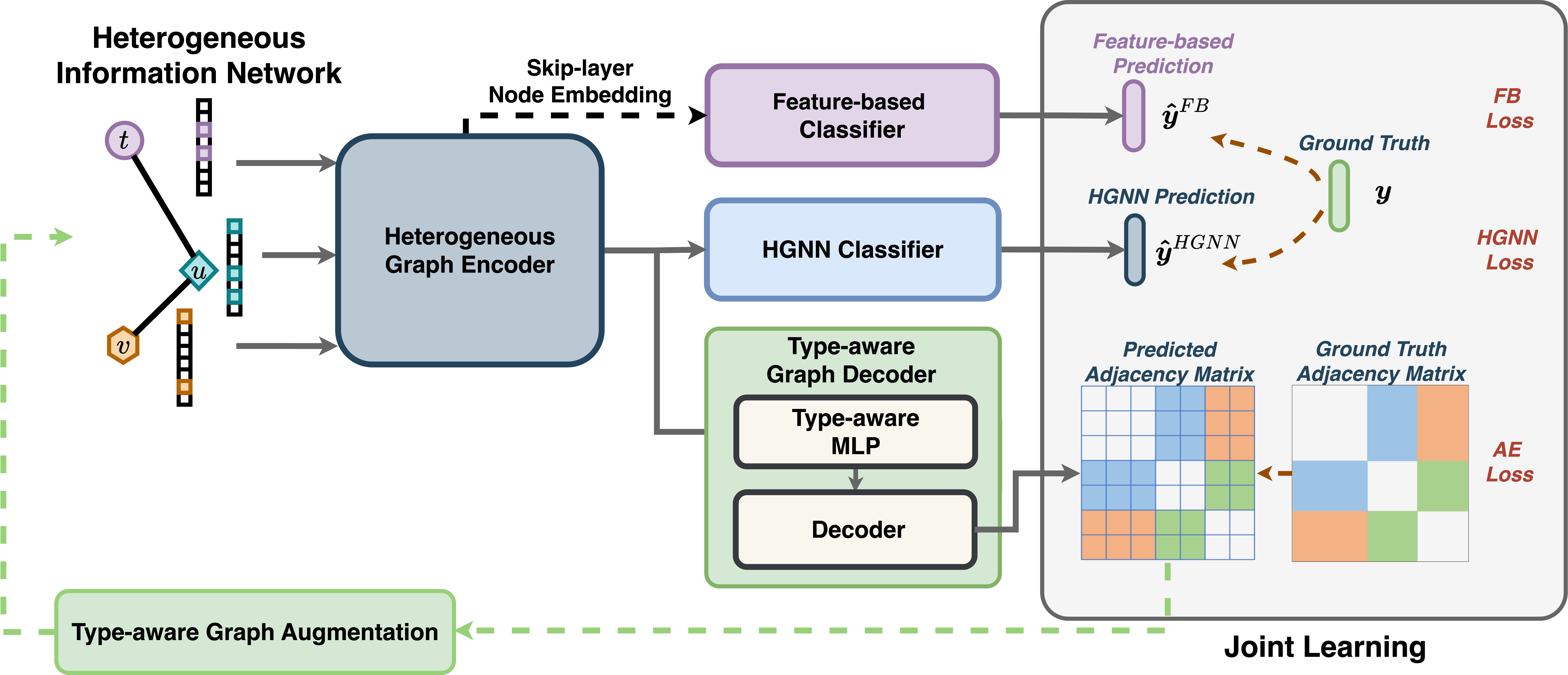}
\caption{The overall framework of THeGAU.}
\label{THeGAE}
\end{figure*}

\section{Proposed Method: THeGAU}\label{sec:method}

\subsection{Preliminaries} 

Heterogeneous information networks, also called heterogeneous graphs, consist of multiple types of nodes and edges, which can be defined as a graph $ G = \{ V, E, \phi, \psi \} $ where $V$ is the set of nodes, and $E$ is the set of edges. 
Each node $v$ has a type $\phi(v)$, and each edge $e$ has a type $\psi(e)$. 
The sets of possible node types and edge types are denoted by $T_v = \{\phi(v) \mid \forall v \in V\}$ and $T_e = \{\psi(e) \mid \forall e \in E\}$. The number of node types and edge types is $|T_v|$ and $|T_e|$, respectively. For a HIN, $|T_v| + |T_e| > 2$.
The node feature can be collected as a feature vector $X$,  where the size of the $X$ is determined by the number of nodes $N$, and node types determine the feature dimension.
Matrix $A$ is the adjacency matrix describing the connection between nodes. 

\subsection{Type-Aware Heterogeneous Graph AutoEncoder and Augmentation Framework}

The THeGAU framework consists of five modules: (1) Heterogeneous Graph Encoder, (2) HGNN Classifier (HGC), (3) Type-aware Graph Decoder (TGD), (4) Feature-based Classifier (FBC), and (5) Type-aware Graph Augmentation (TG-Aug), as shown in Figure~\ref{THeGAE}. The overall process of the THeGAU framework is described in Algorithm~\ref{alg:alg1}.

\begin{algorithm}[t]
\caption{THeGAU Training with Type-Aware Augmentation}\label{alg:alg1}
\begin{algorithmic}[1]

\STATE \textbf{Input:} Heterogeneous graph dataset $\mathcal{G} = (V, E, X, \phi, \psi)$
\STATE \hspace{0.5cm} $\triangleright$ Node features $X$, type mappings $\phi$, $\psi$
\STATE \hspace{0.5cm} Train/validation/test splits: $train\_idx$, $valid\_idx$, $test\_idx$

\STATE \textbf{Initialize:} THeGAU modules (HGNN, FBC, HGC, TGD), dimension $d$, weights $\alpha, \beta$
\STATE Loss composition: $\mathcal{L} = \alpha \mathcal{L}_{AE} + \beta \mathcal{L}_{FB} + (1 - \alpha - \beta) \mathcal{L}_{HGNN}$

\STATE \textbf{Training Loop:}
\FOR{epoch in $[1, \dots, T]$}
    \STATE \textbf{Forward Pass:}
    \STATE \hspace{0.5cm} $Z, H_s = \textbf{HGNN}(X, A, \phi, \psi)[train\_idx]$
    \STATE \hspace{0.5cm} $\hat{Y}^{HGNN} = \textbf{HGC}(Z)$
    \STATE \hspace{0.5cm} $\hat{Y}^{FB} = \textbf{FBC}(H_s)$
    \STATE \hspace{0.5cm} $A_{\text{legal}} = \texttt{Legal\_edge}(G, \phi, \psi)$
    \STATE \hspace{0.5cm} $\hat{A}_{\text{legal}} = \textbf{TGD}(Z, A_{\text{legal}})$

    \STATE \textbf{Compute Losses:}
    \STATE \hspace{0.5cm} $\mathcal{L}_{HGNN} = \texttt{CrossEntropy}(Y, \hat{Y}^{HGNN})$
    \STATE \hspace{0.5cm} $\mathcal{L}_{FB} = \texttt{CrossEntropy}(Y, \hat{Y}^{FB})$
    \STATE \hspace{0.5cm} $\mathcal{L}_{AE} = \frac{1}{|A_{\text{legal}}|} \sum_{(u,v) \in A_{\text{legal}}} \texttt{Focal\_loss}(a_{u,v}, \hat{a}_{u,v})$

    \STATE \textbf{Backward:}
    \STATE \hspace{0.5cm} $\texttt{optimizer.zero\_grad}()$
    \STATE \hspace{0.5cm} $\mathcal{L}.backward()$
    \STATE \hspace{0.5cm} $\texttt{optimizer.step}()$

    \STATE \textbf{Validation (No Gradient):}
    \STATE \hspace{0.5cm} Evaluate $\hat{Y}_{valid}^{HGNN}, \hat{Y}_{valid}^{FB}$ on $valid\_idx$
    \STATE \hspace{0.5cm} Update best model if validation improves; otherwise increment early stop counter
\ENDFOR

\STATE \textbf{Type-Aware Graph Augmentation:}
\STATE \hspace{0.5cm} $A_{\text{aug}, \psi} = \texttt{GraphAugment}(\hat{A}_{\text{legal}}, \texttt{thr}_{add}, \texttt{thr}_{rm}, \psi)$
\STATE \hspace{0.5cm} $G_{\text{aug}} = (X, A_{\text{aug}, \psi}, \phi, \psi)$
\STATE \hspace{0.5cm} Retrain using $G_{\text{aug}}$ and repeat training loop

\STATE \textbf{Inference:}
\STATE \hspace{0.5cm} Evaluate on $test\_idx$ using the best model
\STATE \hspace{0.5cm} Compute final performance using $\hat{Y}^{HGNN}_{test}$
\end{algorithmic}
\label{alg:thegau}
\end{algorithm}

THeGAU is a model-agnostic framework compatible with most HGNNs, designed to enhance graph representation learning and augmentation. THeGAU employs a joint learning approach for end-to-end semi-supervised training, where the HGC handles the primary task, and the TGD and FBC serve as auxiliary tasks. The TGD predicts connections between nodes, acting as an edge predictor for model-based graph augmentation (TG-Aug). This process refines the original graph by removing unnecessary edges and adding beneficial ones, thereby improving message passing. By incorporating TGD as auxiliary losses, the framework preserves essential graph structure information, improving HGNN training and prediction accuracy.
Additionally, FBC employs a skip-layer design to directly connect the feature projection layer to the classifier, effectively addressing over-smoothing and information compression issues while ensuring efficient feature extraction. Detailed descriptions of THeGAU are provided in the following subsections.

\subsection{Heterogeneous Graph Encoder}\label{ssec:hgnn}

As a model-agnostic framework, THeGAU can be seamlessly integrated with a variety of message-passing Heterogeneous Graph Neural Networks (HGNNs). To demonstrate its generality and effectiveness, we instantiate THeGAU with four representative HGNN backbones: HGT \cite{Hu2020HGT}, SimpleHGN \cite{Lv2021SimpleHGN}, SlotGAT \cite{slotgat}, and TreeXGNN \cite{10096251}. These models span a diverse range of architectural designs, showcasing THeGAU’s broad applicability across heterogeneous graph encoding paradigms.
A general HGNN can be expressed as:
\begin{equation}
    Z =  HGNN(X, A, \phi, \psi), 
\end{equation}
where $Z$ denotes the output node embeddings generated by the HGNN, $X$ is the node feature set, and $A$ is the adjacency matrix. 
The embeddings $Z$ serve as the inputs for the heterogeneous graph classifier and type-aware graph decoder. 

\subsection{Type-Aware Graph Decoder (TGD)}\label{tagdm}

The motivation of TGD is to preserve structural semantics and node-type distinctions during edge reconstruction. In homogeneous graph autoencoders such as VGAE, a shared decoder is applied to all node pairs, collapsing heterogeneous semantics into a unified latent space and leading to semantic homogenization.

To address this, THeGAU applies independent MLPs to node embeddings based on their types, generating type-aware representations. These are then used to predict schema-valid edges, enabling reconstruction that respects the graph schema and maintains heterogeneity.

\begin{figure}[h]
  \centering
  \includegraphics[width=\columnwidth]{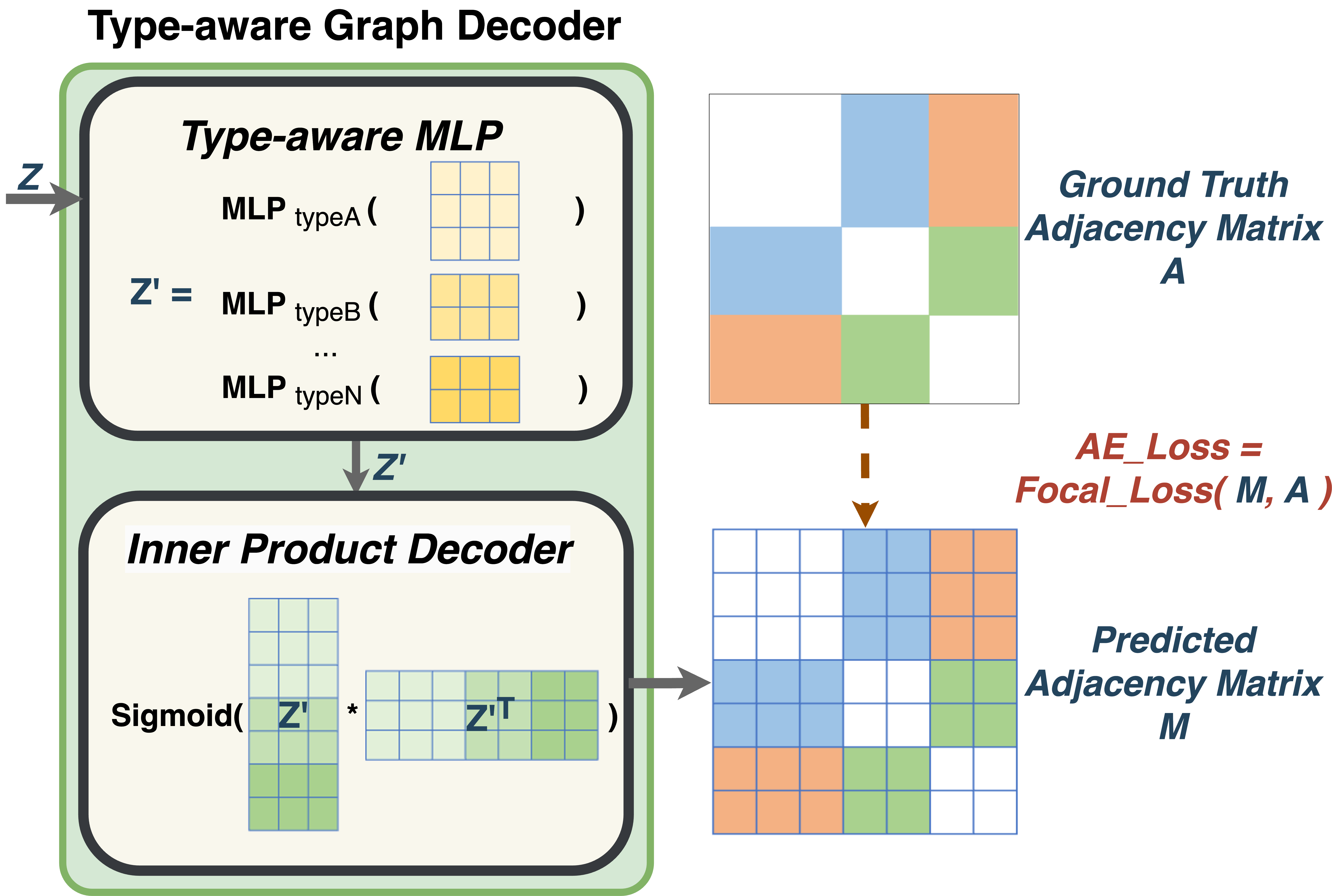}
  \caption{Type-aware Graph Decoder: Node embeddings $Z$ are split by type and processed through multiple MLPs to produce updated embeddings $Z^{\prime}$. An inner product decoder computes similarities to predict edges for valid types, with focal loss applied to measure the error between predictions and ground truth.}
  \label{typeaware} 
\end{figure}

The resulting TGD serves as an auxiliary task that guides the model in reconstructing typed connections and preserving the heterogeneous structure. As illustrated in Figure~\ref{typeaware}, TGD takes the HGNN output embeddings and predicts a type-aware adjacency matrix, helping the model retain meaningful relational patterns in the learned representations.

\begin{figure}[h]
  \centering
  \includegraphics[width=\columnwidth]{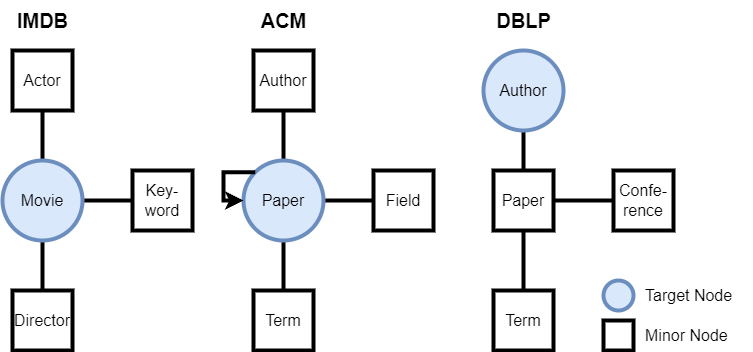}
  \caption{Graph schema of the three heterogeneous graph datasets. Blue nodes represent the target node type, and black box nodes represent the minor node types. Edges between nodes represent existing edge types.}
  \label{3dataset} 
\end{figure}

To enhance the ability to handle the complexity of HINs, we begin by scanning the graph schema, an abstract representation of the graph that defines node and edge types as well as their relationships, as shown in Figure~\ref{3dataset}. Based on this schema, we define a $\texttt{Legal\_edge}$ filter to extract only legal edges $A_{\text{legal}}$, which represent the edges that the model should predict. For example, in IMDB\footnote{\url{https://www.kaggle.com/karrrimba/movie-metadatacsv}}, there are only three legal edge types: [Movie-Actor], [Movie-Keyword], and [Movie-Director]. These are defined as legal edges, while edge types such as [Actor-Keyword], [Actor-Director], and [Keyword-Director] are filtered out. This approach ensures that model training and prediction focus exclusively on the valid and meaningful edge types within the HIN.
\begin{equation}
    A_{\text{legal}} = \texttt{Legal\_edge}(V, \phi, \psi). 
\end{equation}

For each legal edge connecting two nodes $u$ and $v$, we put them into respective MLPs, named the function $\texttt{MLP\_type}$, and generate the corresponding embeddings $Z^{\prime}_u$ and $Z^{\prime}_v$ with the same dimension $d$. $\texttt{MLP\_type}$ consists of multiple one-layer type-aware MLPs with dimension $d$ and a $\texttt{ReLU}$ activation function.
\begin{equation}
\begin{split}
    Z^{\prime}_u = \texttt{MLP\_type}( Z_u, \phi(u) ),\\
    Z^{\prime}_v = \texttt{MLP\_type}( Z_v, \phi(v) ).\\
\end{split}
\end{equation}

We utilize the inner product decoder to predict the edge probabilities as follows: 
\begin{equation}
\begin{split}
    & \hat{a}_{u,v} = \sigma(Z^{\prime}_u * Z^{\prime}_v),\\
\end{split}
\end{equation}
where $*$ indicates the inner product, $\sigma$ is a sigmoid function. We store only the valid edge types instead of the entire $N \times N$ adjacency matrix. This design minimizes the computations needed for edge predictions, significantly reducing both storage and computational complexity. 

The adjacency matrix under consideration is typically sparse, leading to severe class imbalance when predicting edge connections. To address this issue, we adopt the \emph{focal loss}~\cite{Lin2017FocalLF}, which down-weights easy examples and focuses training on harder ones. 
The focal loss is formally defined as:
\begin{equation}
\texttt{Focal\_loss}(a, p) = -\tau_a (1 - p_t)^\gamma \log(p_t),
\end{equation}
where \( a \in \{0,1\} \) denotes the ground-truth adjacency label, and \( p \in [0,1] \) is the predicted edge existence probability. We define:
\[
p_t = p \cdot \mathbb{I}(a = 1) + (1 - p) \cdot \mathbb{I}(a = 0),
\quad
\tau_a = \tau_+ \mathbb{I}(a = 1) + \tau_- \mathbb{I}(a = 0),
\]
with \( \mathbb{I}(\cdot) \) denoting the indicator function. The focusing parameter \( \gamma \geq 0 \) modulates the loss contribution of well-classified examples. It recovers the standard binary cross-entropy when \( \gamma = 0 \). The weighting factor \( \tau_a \) adjusts for imbalance between positive and negative edges.

We apply focal loss to train the type-aware decoder by averaging over all schema-valid node pairs:
\begin{equation}
\mathcal{L}_{AE} = 
\frac{1}{|A_{\text{legal}}|} 
\sum_{(u,v) \in A_{\text{legal}}} 
\texttt{Focal\_loss}(a_{u,v}, \hat{a}_{u,v}),
\end{equation}
where \( \hat{a}_{u,v} \) is the predicted edge probability between nodes \( u \) and \( v \), and \( a_{u,v} \in \{0,1\} \) is the corresponding ground-truth label.

\subsection{Target Node Feature-Based Classifier (FBC)}

Neural networks typically require multiple layers for effective information extraction, but excessive GNN layers can lead to over-smoothing and over-mixing, reducing performance \cite{Rong2020DropEdge, Chen2019MeasuringAR}.
Techniques such as skip connections, graph normalization, and random dropping have been shown to enhance the training of deep GNNs \cite{chen2022bag}, with skip connections being particularly crucial.
In our framework, the FBC employs a skipped connection design to facilitate training of the shared feature space.  
$\texttt{FB\_classifier}$ utilizes the embedding generated by the projection layer in the heterogeneous graph encoder, as expressed by the following equation:
\begin{equation}
    \hat{\boldsymbol{y}}_{\boldsymbol{u}}^{FB} = \texttt{FB\_classifier}(H_{s}),
\end{equation}
where $\texttt{FB\_classifier}$ is a single-layer classifier, and $H_{s}$ is the shared feature space embedding of the target node from the HGNN projection layer \cite{Lv2021SimpleHGN, 10096251}. The loss of the $\texttt{FB\_classifier}$ is $L_{FB}$ and follows the same setup as the $\texttt{HGNN\_classifier}$. Further details are provided in the next subsection.

\subsection{Heterogeneous Graph Classifier (HGC)}\label{ssec:hgclr}

THeGAU leverages common message-passing HGNNs
as heterogeneous graph encoders and connected to HGC for node classification prediction. 
We design a multilayer perceptron with two layers and an activation function, similar to SimpleHGN \cite{Lv2021SimpleHGN} and TreeXGNN \cite{10096251}, referred to as $\texttt{HGNN\_classifier}$. The prediction $\hat{\boldsymbol{y}}_{\boldsymbol{u}}^{HGNN}$ is expressed as follows: 
\begin{equation}
    \hat{\boldsymbol{y}}_{\boldsymbol{u}}^{HGNN} = \texttt{HGNN\_classifier}(Z_u),
\end{equation}
where $Z_u$ represents the node embedding obtained from the heterogeneous graph encoder, and  
$\mathcal{L}_{HGNN}$ is the loss function of the $\texttt{HGNN\_classifier}$, which varies depending on the node classification task. For multi-class prediction and multi-label node classification, we use the common cross-entropy (CE) and binary cross-entropy (BCE), respectively. 

\subsection{Joint Learning}

To enable an efficient end-to-end semi-supervised training process, we integrate a primary loss, \(\mathcal{L}_{HGNN}\), with two auxiliary losses - namely, the type-aware decoder loss \(\mathcal{L}_{AE}\) and the feature-based loss \(\mathcal{L}_{FB}\). The total loss ($\mathcal{L}$) is defined as follows: 
\begin{equation}
    \mathcal{L} = \alpha \mathcal{L}_{AE} + \beta \mathcal{L}_{FB} + (1 - \alpha - \beta) \mathcal{L}_{HGNN},
\end{equation}
where $\alpha$ and $\beta$ are weighting coefficients constrained to $ 0 < \alpha, \beta < 1 $ with $ \alpha + \beta < 1 $, used to balance the contributions of different loss terms during training.

\subsection{Type-Aware Graph Augmentation (TG-Aug)}\label{tgaug}

Recent advancements have enabled homogeneous graph data augmentation via node and edge manipulations \cite{Rong2020DropEdge, Chen2019MeasuringAR, Zhao2020DataAF}. 
Our research extends this idea by applying TGD to refine heterogeneous graph structures. We improve model prediction accuracy by selectively removing extraneous edges and adding critical ones, thereby optimizing the graph's structural integrity.
We leverage well-trained TGD and estimate the probability of typed edges between nodes instead of random sampling. The high-probability edges are added to the augmented graph, while the low-probability ones are removed. The thresholds for these actions are controlled by hyperparameters $\texttt{thr}_{add}$ and $\texttt{thr}_{rm}$. This can enhance the signal-to-noise ratio in the augmented graph and directly improve the effectiveness of graph augmentation. Specifically, the graph augmentation function is defined as follows: 
\begin{equation}
    A_{\text{aug}, \psi} = \texttt{GraphAugment}( \hat{A}_{\text{legal}}, \texttt{thr}_{add}, \texttt{thr}_{rm}, \psi ),
\end{equation}
where $A_{\text{aug}, \psi}$ is the augmented type-aware adjacency matrix, $\hat{A}_{\text{legal}}$ is the predicted edge probability set from Subsection~\ref{tagdm}, and $\psi$ identifies edge types. The hyperparameters $\texttt{thr}_{add}$ and $\texttt{thr}_{rm}$ define thresholds for adding and removing edges, respectively, and are applied separately to each edge type. If a predicted edge’s probability exceeds the corresponding $\texttt{thr}_{add}$, it is added to enhance message passing; if it falls below the corresponding $\texttt{thr}_{rm}$, it is removed to reduce noise. We ensure $\texttt{thr}_{add} > \texttt{thr}_{rm}$ to avoid conflicting operations. The augmented heterogeneous graph, $G_{\text{aug}}$, is defined as follows:
\begin{equation}
    G_{\text{aug}} = \{ X, A_{\text{aug}, \psi}, \phi, \psi \}.
\end{equation}
A grid search technique was utilized to find the optimal threshold for edge probability based on validation outcomes. 

\section{Experimental Settings} \label{sec:Experiments} 

\textbf{Datasets. }
We conduct experiments on three well-known HIN node classification datasets: IMDB, ACM \cite{Wang2019}, and DBLP\footnote{\url{http://web.cs.ucla.edu/~yzsun/data/}}. 
Following the transductive learning setting adopted by prior works \cite{Wang2019, Lv2021SimpleHGN, slotgat, 10096251}, we randomly sample 24\% of the target nodes for training, 6\% for validation, and the remaining 70\% for testing across all three datasets. The statistics of the three datasets are shown in Table~\ref{3Statistics}, including the number of node and edge types, the number of nodes and edges, the number of target node features, and the number of classes.
\begin{table*}[t]
\caption{Statistics of three heterogeneous graph datasets.}
\label{3Statistics}
\centering
\begin{tabular}{lccccccc}
\toprule
 & \multirow{2}{*}[-2pt]{\#Nodes} & \multirow{2}{*}[-2pt]{\#Node Types} 
 & \multirow{2}{*}[-2pt]{\#Edges} & \multirow{2}{*}[-2pt]{\#Edge Types} 
 & \multicolumn{2}{c}{Target Node} & \multirow{2}{*}[-
 2pt]{\#Classes}\\
 \cmidrule(r){6-7}
 &&&&& Type & \#Features &\\
\midrule
 IMDB &19,933 &4 &80,682 &6 &Movie &3,489 &5\\
 ACM & 10,967 & 4 & 551,970 & 8 & Paper & 1,902 & 3\\
 DBLP & 27,303 & 4 & 296,492 & 6 & Author & 340 & 4\\
\bottomrule
\end{tabular}
\end{table*}
\newline
\textbf{Evaluation Metrics.}
We use Micro-F1 and Macro-F1 as evaluation metrics. All experiments were repeated five times, and the average accuracy and standard deviation were computed to assess both performance and stability. Details can be found in the Appendix~\ref{evaluations}.
\newline
\textbf{Baselines. }
We evaluate THeGAU performance via a comprehensive comparison with state-of-the-art HGNNs, including HAN \cite{Wang2019}, DisenHAN \cite{DisenHAN}, GTN \cite{Yun2019GTN}, RSHN \cite{RSHN}, HetGNN \cite{HetGNN}, MAGNN \cite{Fu2020}, HetSANN \cite{HetSANN}, HGT \cite{Hu2020HGT}, SimpleHGN \cite{Lv2021SimpleHGN}, SlotGAT \cite{slotgat}, and TreeXGNN \cite{10096251}. A brief introduction of baselines can be found in the Appendixx~\ref{baselines}.
\newline
\textbf{Environments and Parameter Settings. }
Details of the environments and parameter settings are provided in the Appendix \ref{env_par}, and the model parameters are outlined in Table~\ref{parameter} of Appendix~\ref{hyperparameters}.

\section{Results and Discussion}

\subsection{Performance of the THeGAU}

We evaluated the THeGAU framework using three common HIN datasets. The results, as shown in Table~\ref{Performance}, highlight its effectiveness, with significant improvements in predictive accuracy. We apply THeGAU to four famous HGNNs and observe the improvements. Models with TG- refer to the enhanced HGNNs utilizing our THeGAU framework. The performance of all four HGNN models improved with the addition of THeGAU, with reduced standard deviations in most cases, demonstrating THeGAU's ability to enhance and stabilize various HGNNs. 
Notably, TG-TreeXGNN (Our) and TG-SimpleHGN (Our) achieved statistically significant improvements. TG-SimpleHGN demonstrated remarkable gains, with a 3.4\% increase on IMDB, 2.4\% on ACM, and 1.1\% on DBLP in Macro-F1. Similarly, TG-TreeXGNN achieved a 1.5\% improvement on IMDB and approximately 0.5\% on ACM and DBLP. 
We believe that both models, designed to use a shared feature space, allow TGD and FBC auxiliary tasks to effectively assist in the training of HGNN.
TG-HGT also showed significant improvements in the IMDB and ACM datasets. However, TG-SlotGAT(Our) exhibited only minor improvements, likely due to its slot-based message-passing design, which preserves distinct semantics across feature spaces. This design inherently limits the model's degrees of freedom, making it more challenging to capture additional structural information.

\begin{table*}[t]
\centering
\caption{Performance comparison. Models with TG- refer to the enhanced HGNNs utilizing our THeGAU framework. We use a bold font to highlight the model with the THeGAU framework with performance improvement and denote significance levels with * for p-value $<$ 0.05, ** for p-value $<$ 0.01, and *** for p-value $<$ 0.001.}
{
\small
\begin{tabular}{lllllll}
\toprule
 & \multicolumn{2}{c}{IMDB}
 & \multicolumn{2}{c}{ACM}
 & \multicolumn{2}{c}{DBLP}\\
\cmidrule(r){2-3} \cmidrule(r){4-5} \cmidrule(r){6-7}
 & Macro-F1 (\%) & Micro-F1 (\%)
 & Macro-F1 (\%) & Micro-F1 (\%)
 & Macro-F1 (\%) & Micro-F1 (\%)\\
\midrule
 HAN \cite{Wang2019} & 57.74±0.96 & 64.63±0.58
 &90.89±0.43 & 90.79±0.43
 &91.67±0.49 & 92.05±0.62\\
 DisenHAN \cite{DisenHAN} & 63.40±0.49 & 67.48±0.45
 &92.52±0.33 & 92.45±0.33
 &93.66±0.39 & 94.18±0.36\\
 GTN \cite{Yun2019GTN} & 60.47±0.98 & 65.14±0.45
 & 91.31±0.70 & 91.20±0.71
 & 93.52±0.55 & 93.97±0.54\\
 RSHN \cite{RSHN} & 59.85±3.21 & 64.22±1.03
 & 90.50±1.51 & 90.32±1.54
 & 93.34±0.58 & 93.81±0.55\\
 HetGNN \cite{HetGNN} & 48.25±0.67 & 51.16±0.65
 & 85.91±0.25 & 86.05±0.25
 & 91.76±0.43 & 92.33±0.41\\
 MAGNN \cite{Fu2020} & 56.49±3.20 & 64.67±1.67
 & 90.88±0.64 & 90.77±0.65
 & 93.28±0.51 & 93.76±0.45\\
 HetSANN \cite{HetSANN} & 49.47±1.21 & 57.68±0.44
 & 90.02±0.35 & 89.91±0.37
 & 78.55±2.42 & 80.56±1.50\\
\midrule
 HGT \cite{Hu2020HGT} & 62.70±0.79 & 65.34±0.80
 & 92.62±0.34 & 92.58±0.34
 & 93.02±0.44 & 93.58±0.41\\
 \textbf{TG-HGT (Our)} & \textbf{63.92±1.08} & \textbf{67.91±0.72\textsuperscript{***}}
 & \textbf{93.90±0.30\textsuperscript{***}} & \textbf{93.86±0.29\textsuperscript{***}}
 & \textbf{93.25±0.26} & \textbf{93.82±0.24}\\
\midrule
 SimpleHGN \cite{Lv2021SimpleHGN} & 63.50±0.66 & 66.31±0.49
 & 92.32±0.55 & 92.29±0.55
 & 93.42±0.36 & 93.87±0.37\\
\textbf{TG-SimpleHGN (Our)} & \textbf{65.69±0.13\textsuperscript{***}} & \textbf{68.01±0.14\textsuperscript{***}}
 & \textbf{94.49±0.32\textsuperscript{***}} & \textbf{94.45±0.34\textsuperscript{***}}
 & \textbf{94.42±0.18\textsuperscript{***}} & \textbf{94.87±0.16\textsuperscript{***}}\\
\midrule
 SlotGAT \cite{slotgat} & 63.45±1.09 & 68.43±0.44
 & 93.91±0.39 & 93.83±0.40
 & 94.69±0.05 & 95.07±0.06\\
\textbf{TG-SlotGAT (Our)} & \textbf{64.57±0.50} & \textbf{68.69±0.24}
 & \textbf{94.08±0.16} & \textbf{94.00±0.16}
 & \textbf{94.79±0.28} & \textbf{95.18±0.22}\\
\midrule
 TreeXGNN \cite{10096251} & 64.90±0.54 & 68.94±0.27
 & 94.44±0.46 & 94.42±0.45
 & 94.14±0.71 & 94.59±0.66\\
 \textbf{TG-TreeXGNN (Our)} & \textbf{65.85±0.23\textsuperscript{**}} & \textbf{69.64±0.19\textsuperscript{**}}
 & \textbf{94.79±0.27\textsuperscript{*}} & \textbf{94.75±0.27\textsuperscript{*}}
 & \textbf{94.63±0.06\textsuperscript{*}} & \textbf{95.02±0.07\textsuperscript{*}}\\
\bottomrule
\end{tabular}
}
\label{Performance}
\end{table*}

\begin{table*}[t]
\centering
\caption{Ablation studies of THeGAU. We removed components of the THeGAU individually to verify the importance of each element. ``w/o” indicates that the corresponding component is removed. We compare the ablation variants against the full model and denote significance levels with * for p-value $<$ 0.05, ** for p-value $<$ 0.01, and *** for p-value $<$ 0.001.}
{
\small
\begin{tabular}{lllllll}
\toprule
 & \multicolumn{2}{c}{IMDB}
 & \multicolumn{2}{c}{ACM}
 & \multicolumn{2}{c}{DBLP}\\
\cmidrule(r){2-3} \cmidrule(r){4-5} \cmidrule(r){6-7}
 & Macro-F1 (\%) & Micro-F1 (\%)
 & Macro-F1 (\%) & Micro-F1 (\%)
 & Macro-F1 (\%) & Micro-F1 (\%)\\
\midrule
 \textbf{TG-SimpleHGN} 
 & \textbf{65.69±0.13} & \textbf{68.01±0.14}
 & \textbf{94.49±0.32} & \textbf{94.45±0.34}
 & \textbf{94.42±0.18} & \textbf{94.87±0.16}\\
 (w/o TG-Aug) 
 & 65.49±0.17 & 67.63±0.31\textsuperscript{*}
 & 94.40±0.43 & 94.36±0.44
 & 94.31±0.41 & 94.76±0.39\\
 (w/o TG-Aug \& TGD-MLP) 
 & 64.68±0.71\textsuperscript{*} & 66.61±0.81\textsuperscript{**}
 & 93.52±0.84\textsuperscript{*} & 93.49±0.87
 & 94.03±1.58 & 94.53±1.40\\
 (w/o TG-Aug \& TGD)
 & 63.01±1.19\textsuperscript{**} & 67.16±0.14\textsuperscript{***}
 & 93.36±0.89\textsuperscript{*} & 93.31±0.90\textsuperscript{*}
 & 93.65±0.66\textsuperscript{*} & 94.14±0.62\textsuperscript{*}\\
 (w/o TG-Aug \& FBC)
 & 63.69±0.61 & 67.62±0.57
 & 93.63±0.99 & 93.58±1.03
 & 93.81±0.75 & 94.32±0.68\\
\midrule
 \textbf{TG-TreeXGNN} 
 & \textbf{65.85±0.23} & \textbf{69.64±0.19}
 & \textbf{94.79±0.27} & \textbf{94.75±0.27}
 & \textbf{94.63±0.06} & \textbf{95.02±0.07}\\
 (w/o TG-Aug) 
 & 65.58±0.38 & 69.64±0.19
 & 94.66±0.25 & 94.62±0.26
 & 94.55±0.13 & 94.94±0.11\\
 (w/o TG-Aug \& TGD-MLP)
 & 63.26±1.85\textsuperscript{*} & 67.71±1.59\textsuperscript{**}
 & 93.66±0.72\textsuperscript{*} & 93.63±0.69\textsuperscript{**}
 & 94.39±0.33 & 94.80±0.27\\
 (w/o TG-Aug \& TGD)
 & 65.55±0.35 & 67.60±0.80\textsuperscript{***}
 & 94.21±0.43\textsuperscript{*} & 94.18±0.43\textsuperscript{*}
 & 94.25±0.06\textsuperscript{***} & 94.70±0.05\textsuperscript{***}\\
 (w/o TG-Aug \& FBC)
 & 65.51±0.54 & 69.00±0.39\textsuperscript{*}
 & 93.58±0.81\textsuperscript{*} & 93.55±0.82\textsuperscript{*}
 & 94.22±1.11 & 94.65±1.01\\
\bottomrule
\end{tabular}
}
\label{ablation}
\end{table*}

\subsection{Ablation Studies}\label{sec_ablation}

To assess the contribution of each module in THeGAU, we conducted an ablation study by systematically removing critical modules one at a time. This analysis focused on two HGNNs, SimpleHGN and TreeXGNN, as representative examples. The experimental results are summarized in Table~\ref{ablation}. The abbreviations ``TG-Aug,'' ``TGD-MLP,'' ``TGD," and ``FBC'' refer to the type-aware graph augmentation module, the MLPs in the TGD module, the entire TGD module, and the FBC module in the THeGAU framework, respectively. 
We observed that models with the complete THeGAU outperformed others by approximately 1\% to 3\% in average Macro-F1, with a significant improvement in model stability (the standard deviation decreased in 23 out of 24 cases), highlighting the effectiveness of the THeGAU design.

Compared with THeGAU, ``w/o TG-Aug'' has lower prediction accuracy consistently, with an average Macro/Micro-F1 score decrease observed in 11 out of 12 cases. Additionally, in three-quarters of the cases, the standard deviation has increased, indicating worse stability. This proves the effectiveness of type-aware graph augmentation. TGD serves effectively as an edge predictor for model-based graph augmentation, aiding in noise reduction and optimizing message transmission within real-world HINs. 

Compared with the ``w/o TG-Aug,'' ``w/o TG-Aug \& TGD-MLP'' has significantly lower model accuracy and stability, especially for TG-TreeXGNN, which dropped 3.6\% in Macro-F1 for IMDB. It is evident that the design of TGD-MLP plays a crucial role in the TG-TreeXGNN. 

TG-SimpleHGN ``w/o TG-Aug \& TGD'' also shows a noticeable performance drop, which is around a 4\% decrease in Macro-F1 for the IMDB dataset. Therefore, TGD is a critical module for TG-SimpleHGN. We observe that differences exist among HGNN encoders and datasets. However, the type-aware graph decoder with an MLP design consistently performs stably and delivers the best results across various models and datasets. This stability highlights its effectiveness in preserving and leveraging the complex information within HINs for enhanced performance.

The ``w/o TG-Aug \& FBC" model shows significantly lower accuracy and stability than ``w/o TG-Aug," with standard deviations roughly twice as high across two models and three datasets. This underscores the effectiveness of FBC, a skip-connection design. Ablation studies highlight the critical role of each module in the THeGAE framework.

\subsection{Design of the Type-Aware Graph Decoder}\label{ssec:hgdecoderva}

\begin{figure}[h]
\centering
\includegraphics[width=\columnwidth]{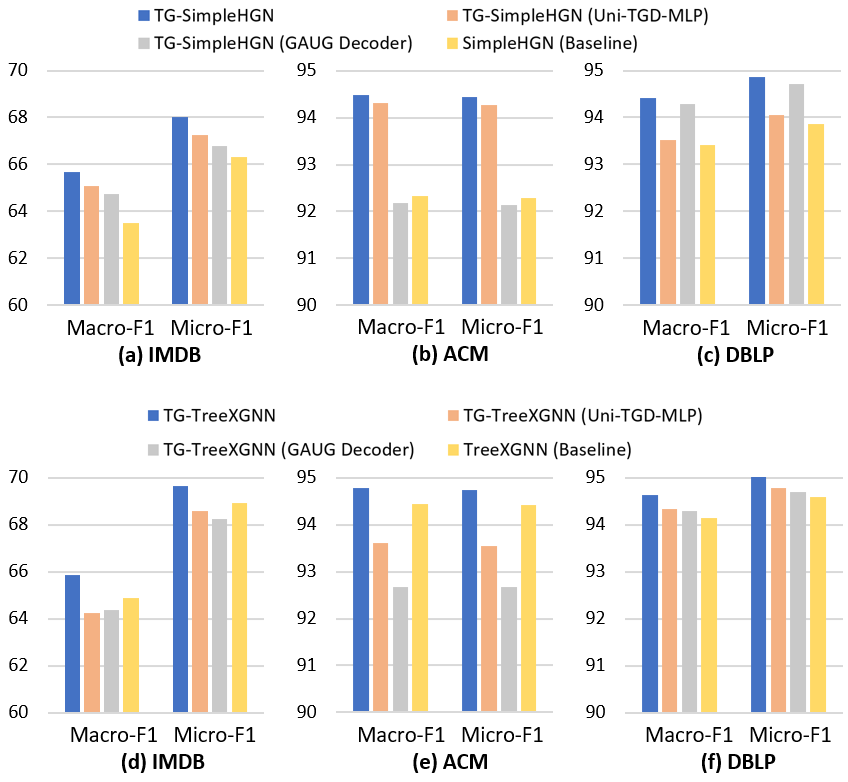}
\caption{
Type-aware Graph Decoder Variations. TG-SimpleHGN and TG-TreeXGNN are the proposed methods. ``Uni-TGD-MLP'' uses a universal MLP decoder applied uniformly across edge types, while ``GAUG Decoder'' applies the homogeneous graph decoder from GAUG directly.}
\label{TypeAwareDecoderVariation}
\end{figure}

We examined different graph decoder variants: (1) Type-aware decoder we proposed; (2) ``Uni-TGD-MLP'' that applies a single MLP decoder, i.e., the same MLP regardless of edge type; (3) ``GAUG Decoder'' \cite{Zhao2020DataAF}, a homogeneous graph decoder; and (4) the baseline models without THeGAU. 
Figure~\ref{TypeAwareDecoderVariation} shows that TG-SimpleHGN and TG-TreeXGNN outperformed all variants, with the performance order typically being: THeGAU TGD $>$ Uni-TGD-MLP $>$ GAUG Decoder. This highlights the necessity of a well-designed type-aware graph decoder tailored for heterogeneous data to enhance model learning and accuracy.
The Uni-TDG-MLP yields inconsistent gains, highlighting the need for a specialized type-aware graph decoder to handle the complexities of node relationships. A single, uniform decoder like Uni-TGD-MLP is often too simplistic to capture graph nuances effectively, making it an insufficient edge predictor. The GAUG decoder \cite{Zhao2020DataAF}, originally designed for homogeneous graphs, underperforms in heterogeneous graphs and negatively impacts HGNN models like SimpleHGN and TreeXGNN, especially on ACM datasets (Figure~\ref{TypeAwareDecoderVariation} (b), (e)). This underscores the importance of a specialized type-aware architecture combined with an MLP, which can improve adaptability and predictive performance. The proposed TGD shows promising potential for future research.

\subsection{Parameter Sensitivity}

$\alpha$ and $\beta$ are critical hyperparameters in our proposed method. To evaluate their influence on model performance, we conducted sensitivity experiments on the IMDB dataset using TG-SimpleHGN. In the first set of experiments, we fixed $\beta$ and varied $\alpha$ to analyze its effect. Based on the validation performance, we selected the optimal parameters and reported the corresponding test results. 

The validation results indicate that the model achieves peak performance when $\alpha$ is set to 0.3, with a relatively stable high-performance region observed between $\alpha$ = 0.2 and 0.3 on the test set, and the standard deviation is relatively smaller than 0.1 and 0.4 (from 0.31 to 0.62 and 1.05 in Micro-F1, respectively). 

\begin{table}[ht]
\centering
\caption{Parameter sensitivity analysis on the IMDB dataset using TG-SimpleHGN under different $\alpha$ values with fixed $\beta = 0.1$.}
{
\setlength{\tabcolsep}{0.8mm}
\begin{tabular}{ccccc}
\toprule
 & \multicolumn{2}{c}{validation set} & \multicolumn{2}{c}{test set} \\
\cmidrule(lr){2-3} \cmidrule(lr){4-5}
$\alpha$ & Macro-F1 (\%) &Micro-F1 (\%) & Macro-F1 (\%) & Micro-F1 (\%) \\
\midrule
0.1 & 65.06$\pm$2.23 & 66.68$\pm$2.33 & 65.41$\pm$0.55 & 67.48$\pm$0.62 \\
0.2 & 65.67$\pm$2.06 & 67.33$\pm$2.05 & 65.64$\pm$0.12 & 67.65$\pm$0.38 \\
\textbf{0.3} & \textbf{66.90$\pm$1.48} & \textbf{68.46$\pm$1.61} & \textbf{65.49$\pm$0.17} & \textbf{67.63$\pm$0.31} \\
0.4 & 65.75$\pm$1.61 & 67.36$\pm$1.81 & 64.92$\pm$1.00 & 66.83$\pm$1.05 \\
\bottomrule
\end{tabular}
}
\label{tab:alpha-ablation}
\end{table}

Similarly, when $\alpha$ was fixed, and we varied $\beta$, the best performance was observed at $\beta$ = 0.1. As $\beta$ increased from 0.2 to 0.4, the performance exhibited fluctuations and a gradual decline. 

\begin{table}[ht]
\caption{Parameter sensitivity analysis on the IMDB dataset using TG-SimpleHGN under different $\beta$ values with fixed $\alpha = 0.3$.}
\centering
{
\begin{tabular}{ccccc}
\toprule
 & \multicolumn{2}{c}{validation test} & \multicolumn{2}{c}{test set} \\
\cmidrule(lr){2-3} \cmidrule(lr){4-5}
$\beta$ & Macro-F1 (\%) & Micro-F1 (\%) & Macro-F1 (\%) & Micro-F1 (\%) \\
\midrule
\textbf{0.1} & \textbf{66.90$\pm$1.48} & \textbf{68.46$\pm$1.61} & \textbf{65.49$\pm$0.17} & \textbf{67.63$\pm$0.31} \\
0.2 & 64.63$\pm$1.13 & 66.36$\pm$1.15 & 65.15$\pm$0.63 & 67.18$\pm$0.63 \\
0.3 & 65.51$\pm$1.98 & 67.24$\pm$2.23 & 65.30$\pm$0.27 & 67.60$\pm$0.25 \\
0.4 & 65.71$\pm$0.60 & 67.29$\pm$0.50 & 65.30$\pm$0.40 & 67.52$\pm$0.37 \\
\bottomrule
\end{tabular}
}
\label{tab:beta-ablation}
\end{table}

\subsection{Complexity of THeGAU}\label{complexity}

In the THeGAU framework, the computational complexity of the HGNN encoder and FBC is similar to the prior works \cite{Lv2021SimpleHGN, 10096251}, so it is not the focus here. Instead, we focus on the complexity of the TGD.
The complexity of the homogeneous graph decoder depends on the number of nodes. For example, DropEdge \cite{Rong2020DropEdge} randomly removes a portion of graph edges with low complexity but limited effectiveness, while AdaEdge \cite{Chen2019MeasuringAR} modifies edges through a two-stage process based on a confidence measure. Similarly, GAUG \cite{Zhao2020DataAF} predicts edge connection probabilities, requiring additional $O(N^2)$ complexity for matrix manipulation. However, link prediction in heterogeneous graphs differs as not all nodes are necessarily connected. For example, in the IMDB dataset, edges like [Actor-Keyword], [Actor-Director], and [Keyword-Director] are illegal, as mentioned in subsection~\ref{tagdm}. The proposed TGD focuses only on predicting existing edge types, reducing the complexity from $O(N^2)$ to $O(k \cdot n_u \cdot n_v)$, where $k$ is the number of legal edge types and $n_u$ and $n_v$ are the number of nodes connected by these edge types. This type-aware design significantly reduces complexity and noise compared to traditional graph decoders, which are typically extended from homogeneous graphs.

\section{Conclusion}
\label{sec:Conclusion}

In this work, we introduced THeGAU, a type-aware heterogeneous graph autoencoder and augmentation framework for end-to-end semi-supervised learning on HINs. By incorporating typed edge reconstruction as an auxiliary task, THeGAU preserves node-type semantics, stabilizes training, and mitigates overfitting---leading to improved generalization. Its lightweight and modular design also reduces computational overhead, offering practical gains in both memory and runtime efficiency.

Beyond representation learning, THeGAU functions as a structure-aware augmentation module that denoises the input graph by removing spurious edges and selectively reinforcing meaningful ones. This process enhances the signal-to-noise ratio of the graph and improves model robustness. Extensive experiments on benchmark HIN datasets demonstrate that THeGAU consistently boosts performance across diverse HGNN backbones, validating its effectiveness as a plug-and-play framework for robust heterogeneous graph learning.

While THeGAU is effective on mid-scale HINs, it currently faces scalability challenges due to decoder memory overhead. In future work, we aim to extend THeGAU to large-scale graphs through subgraph-based training or sparse decoding strategies. Additionally, although our current focus is on node classification, the architecture naturally generalizes to link prediction, graph classification, and other downstream tasks---opening new directions for broadening its applicability in real-world heterogeneous graph scenarios.

\bibliographystyle{ACM-Reference-Format}
\bibliography{refs}

\appendix
\section{Evaluation Metrics} \label{evaluations}

We employ Micro-F1 and Macro-F1 as evaluation metrics. Micro-F1 is a metric used to evaluate the performance of a machine learning model in a multi-class classification problem. It takes into account the number of true positives (TP), false positives (FP), and false negatives (FN) to provide a global measure of the model's performance. The Micro-F1 score is also known as F1 in general and is calculated using the following formula:

\begin{equation}
    \text{Micro-F1} = \frac{2 \times \text{TP}}{2 \times \text{TP} + \text{FP} + \text{FN}}.
\end{equation}

Macro-F1 calculates the F1 score for each class individually and then computes their unweighted average. In this case, each class contributes equally to the average F1 score. The Macro-F1 score is calculated using the following formula: 

\begin{equation}
    \text{Macro-F1} = \frac{1}{n} \sum_{i=1}^{n} \text{F1}_i,
\end{equation}
where $n$ is the total number of label categories present in the dataset.

Our objective is to measure the performance of different categories without being influenced by varying class ratios among test samples to ensure unbiased evaluation metrics. Therefore, we primarily utilize Macro-F1 as the main evaluation metric, while Micro-F1 serves as an auxiliary metric.
All experiments were repeated five times, and the average accuracy and standard deviation were used to evaluate performance and stability.

\section{Baselines}\label{baselines}

We evaluate the performance of our proposed framework, THeGAU, via a comprehensive comparison with other state-of-the-art HGNNs. The brief introduction of the baseline models is
described as follows:

\begin{itemize}
\item HAN \cite{Wang2019} is designed to handle heterogeneous graphs. It leverages hierarchical attention, including node-level and semantic-level attention, to learn the importance of nodes and meta-paths and generate the node embeddings.

\item DisenHAN \cite{DisenHAN} learns disentangled representations by leveraging meta relations and high-order connectivity. It aggregates aspect-specific features for each user/item and captures collaborative filtering effects. 

\item GTN \cite{Yun2019GTN} identifies valuable connections between different types of nodes and edges, eliminating the need for domain-specific graph preprocessing or predefined meta-paths to perform graph representation learning.

\item RSHN \cite{RSHN} integrates both the graph and its coarsened line graph to embed nodes and edges without needing prior knowledge like metapaths, enhancing message propagation between nodes and edges and improving embedding quality. 

\item HetGNN \cite{HetGNN} uses a random walk with a restart strategy to sample strongly correlated heterogeneous neighbors for each node, grouped by node types, and aggregates feature information from the neighboring nodes.

\item MAGNN \cite{Fu2020} incorporates node content transformation to encapsulate input node attributes, intra-metapath aggregation to incorporate intermediate semantic nodes, and inter-metapath aggregation to combine messages from multiple metapaths.

\item HetSANN \cite{HetSANN} encodes the structural information of HIN without relying on meta-paths, utilizing low-dimensional projections and type-specific graph attention to aggregate multi-relational information.

\item HGT \cite{Hu2020HGT} applies transformer-based design to HINs, node- and edge-type dependent parameters were designed to characterize the heterogeneous attention over each edge and empower HGT to maintain dedicated representations for different types of nodes and edges to model heterogeneity.

\item SimpleHGN \cite{Lv2021SimpleHGN} introduces a strong but simple HGNN model, which utilizes projection layers to project information to the shared feature space and aggregates information with an edge-type attention mechanism. 

\item SlotGAT \cite{slotgat} separates message passing by node type in slots, preserving distinct feature spaces. It utilizes a slot attention mechanism and slot-wise aggregation to generate node embeddings for downstream tasks.

\item TreeXGNN \cite{10096251} effectively extracts node features via gradient-boosted decision trees and integrates community structure information with fusion modules and shared feature space design to boost HGNNs. 
\end{itemize}

We re-implement four widely used baselines models, including HGT, SimpleHGN, SlotGAT, and TreeXGNN, as benchmarks. For the remaining models, we rely on the results reported in their respective papers \cite{Lv2021SimpleHGN, slotgat} for comparison.

\section{Environments and Parameter Settings} \label{env_par}

We perform our experiments on Ubuntu 20.04.5 LTS, CUDA Version 11.7, Torch 1.12.1+cu113 \cite{PyTorch}, and dgl-cu111 0.9.1 \cite{wang2019dgl} for GNN implementations. The machine is equipped with NVIDIA RTX A6000 GPU.
The HGNN models we used in the THeGAU framework follow a similar setting to the original paper \cite{Hu2020HGT, Lv2021SimpleHGN, 10096251}. 
We used AdamW optimizer with a 30-epoch early stop in 300 epochs for training and selected the best based on validation. We applied grid-search on the crucial hyperparameters, such as the hidden dimensions for HGNN encoder and decoder, $d$, within \{32, 64, 128\}, and joint learning hyperparameters $\alpha$ and $\beta$ within \{0.1, 0.2, 0.3\}. The focal loss $\gamma$ is 2, recommended by \cite{Lin2017FocalLF}. The number of attention heads is 8, and the number of HGNN layers is 3. We conduct a grid search for $thr_{add}$ and $thr_{rm}$ values during the refinement process. %We consider a range of parameters within \{1-1e-5, 1-1e-6, 1-1e-7\} and \{1e-5, 1e-6, 1e-7\}, respectively.

\section{Model Parameters} \label{hyperparameters}

Table~\ref{parameter} presents the parameters used in this evaluation. Given the varying characteristics of different models, the corresponding hyperparameters were adjusted accordingly. With appropriate tuning, THeGAU demonstrated excellent performance. These results collectively affirm the flexibility and universal applicability of the THeGAU framework, highlighting its substantial potential to enhance the performance of diverse HGNN encoders across various datasets, shown in Table~\ref{Performance}.

\begin{table}[ht]
\centering
\caption{THeGAU hyperparameters. HGNN hidden dimensions are represented by $d$, and the hyperparameters for joint learning are $\alpha$ and $\beta$ for the type-aware decoder and feature-based loss}
{
\small
\setlength{\tabcolsep}{1mm}
\begin{tabular}{lccccccccc}
\toprule
 & \multicolumn{3}{c}{IMDB} 
 & \multicolumn{3}{c}{ACM} 
 & \multicolumn{3}{c}{DBLP}\\
\cmidrule(lr){2-4} \cmidrule(lr){5-7} \cmidrule(lr){8-10}
 & $d$ & $\alpha$ & $\beta$ 
 & $d$ & $\alpha$ & $\beta$ 
 & $d$ & $\alpha$ & $\beta$\\
\midrule
 TG-HGT & 64 & 0.1 & - 
 & 64 & 0.1 & - 
 & 64 & 0.2 & -\\
 TG-SimpleHGN & 64 & 0.3 & 0.1 
& 32 & 0.3 & 0.1 
& 64 & 0.2 & 0.1\\
 TG-SlotGAT & 128 & 0.1 & 0.0 
& 64 & 0.3 & 0.0 
& 64 & 0.1 & 0.0\\
 % \midrule
 TG-TreeXGNN & 64 & 0.1 & 0.3 
& 64 & 0.1 & 0.3 
& 128 & 0.1 & 0.3\\
\bottomrule
\end{tabular}
}
\label{parameter}
\end{table}

\end{document}